\documentclass[runningheads]{llncs}
\usepackage[T1]{fontenc}
\usepackage{graphicx,arydshln}
\usepackage{hyperref}
\usepackage{color}
\hypersetup{
    colorlinks = true,
    linkcolor = blue,
    anchorcolor = blue,
    citecolor = blue,
    filecolor = blue,
    urlcolor = blue,
    }

\usepackage{xcolor}

\begin{document}
\title{Lymphoid Infiltration Assessment of the Tumor Margins in H\&E Slides}
%
%
\author{Zhuxian Guo\inst{1,2,}\thanks{These authors contributed equally to this work.} \and
Amine Marzouki\inst{1, \star}  \and
Jean-François Emile\inst{3}\and
Henning M\"uller\inst{2}\and
Camille Kurtz\inst{1}\and
Nicolas Loménie\inst{1}}
\authorrunning{Z. GUO, A. MARZOUKI et al.}
%
\institute{Laboratoire d'Informatique Paris Descartes (LIPADE), Université Paris Cité, Paris, France
\and
University of Applied Sciences of Western Switzerland (HES-SO Valais), Sierre, Switzerland
\and
Department of Pathological Anatomy and Cytology, Ambroise-Paré Hospital, Boulogne, France\\
}
\maketitle              
\begin{abstract}

Lymphoid infiltration at tumor margins is a key prognostic marker in solid tumors, playing a crucial role in guiding immunotherapy decisions. Current assessment methods, heavily reliant on immunohistochemistry (IHC), face challenges in tumor margin delineation and are affected by tissue preservation conditions. In contrast, we propose a Hematoxylin and Eosin (H\&E) staining-based approach, underpinned by an advanced lymphocyte segmentation model trained on a public dataset for the precise detection of  CD3$^{+}$ and  CD20$^{+}$ lymphocytes.
In our colorectal cancer study, we demonstrate that our H\&E-based method offers a compelling alternative to traditional IHC, achieving comparable results in many cases. Our method's validity is further explored through a Turing test, involving blinded assessments by a pathologist of anonymized curves from H\&E and IHC slides. This approach invites the medical community to consider Turing tests as a standard for evaluating medical applications involving expert human evaluation, thereby opening new avenues for enhancing cancer management and immunotherapy planning.

\end{abstract}
\section{Introduction}
\label{sec:intro}
In the context of solid tumors, the assessment of lymphoid infiltration plays a crucial role in guiding treatment strategies and determining prognostic outcomes.
Immunohistochemistry (IHC) staining is a conventional method for identifying lymphocytes, yet it suffers from variability due to differing tissue storage conditions, affecting the reliability of lymphocyte quantification. 
Using IHC whole slide imaging (WSI), the linear quantification of lymphoid infiltration (LQLI) \cite{Allard2012LinearFactor} technique, offers a solution by profiling lymphoid infiltration patterns via density variation curves of CD3+ cells within several invasive front segments along a 4-mm-length (i.e., 2.0 mm outside/inside the tumor margin). This method was improved in \cite{DBLP:conf/iciar/DjiroKL18} to evaluate a continuous region along the tumor margin to help identify high immune infiltrate patients with microsatellite stable (MSS) metastatic colorectal cancer (CRC) \cite{GALLOIS20211254}. However, challenges remain in setting precise test cut-off values for infiltration peaks due to IHC's inherent instability. Moreover, the delineation of tumor margins is less precise in IHC compared to Hematoxylin and Eosin (H\&E) stained slides, where cellular structures are more distinct. This has led to a practice among pathologists of evaluating the area under the lymphocyte density peak at tumor margins, focusing on the density surface rather than peak amplitude, to make clinical decisions.

\par

H\&E staining stands as the cornerstone of histology analysis, known for its conventional application and robustness. Importantly, it enables pathologists to delineate tumor margins more intuitively, making it a crucial tool for lymphoid infiltration assessment. This method's potential to facilitate the quantitative evaluation of lymphocyte density peaks at tumor margins is significant, particularly for establishing test cut-off values. 
Herein, we propose a H\&E-based approach for lymphoid infiltration assessment, leveraging a deep learning model trained for precise lymphocyte segmentation, combined with image processing techniques to generate accurate lymphocyte density curves for quantification.

\par

Lymphocyte segmentation is a critical component of our methodology. Utilizing a contextual aware neural network with expanded receptive field, we achieve high accuracy in lymphocyte segmentation from H\&E WSIs. 
This advancement builds on prior work in the field, where small scale datasets were initially used for lymphocyte segmentation. Recent innovations in restaining and co-registration techniques have facilitated the creation of large-scale annotated dataset \cite{DBLP:conf/miccai/GhahremaniMHISCN23}, allowing for enhanced training and validation of our segmentation model. Compared to conventional Fully Convolutional Network (FCN), the contextual aware neural network can effectively learn the cellular context and architecture. The model was meticulously trained on a comprehensive dataset before being externally validated on two additional public datasets, ensuring its efficacy and reliability in lymphoid infiltration assessment across diverse cohorts, before applying on household lymphocyte-label-free cohort.

\par

Our lymphoid infiltration assessment methodology underwent rigorous validation using 12 pairs of household H\&E and IHC slides, demonstrating the concordance of lymphoid infiltration curves derived from H\&E slides with those obtained from IHC slides. This study makes significant methodological advancements in two key areas: (1) the semantic segmentation of lymphocytes from H\&E slides; and (2) the detailed comparison of infiltration curves against established IHC benchmarks, showcasing our method's reliability. From a clinical standpoint, this research paves the way for the integration of this method into routine patient pre-screening for immunotherapy eligibility, offering a substantial contribution to personalized medicine. 

\begin{figure}[t!]
 \centering
  \includegraphics[width=0.63\textwidth]{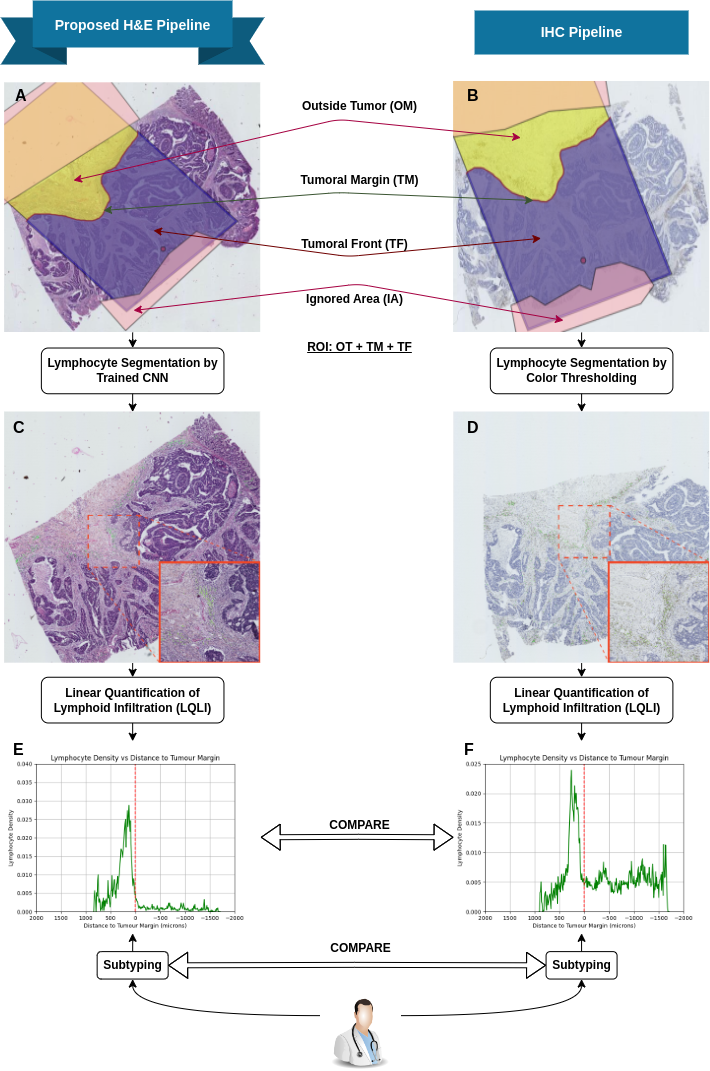}
\caption{Workflow of the lymphoid infiltration assessment of the tumor margin. Annotations by an expert pathologist on the H\&E (A) and its corresponding IHC (B) slides. 
(C), (D): Lymphocyte segmentation results on the paired slides (identified lymphocytes in green). 
Lymphocyte density curves as a function of distance from the tumor margin
from H\&E (E) and IHC (F) slides.}
\label{fig:pochi}
\end{figure}

\section{Datasets}
\label{sec:dataset}
The proposed lymphoid infiltration assessment method is tailored for H\&E slides. 
The obtained results will be validated on a conventional assessment results of IHC slides \cite{DBLP:conf/iciar/DjiroKL18} from the same tumor by assuming the infiltration profiles from adjacent slices remain constant.
The household dataset for tumor margin lymphoid infiltration assessment contains 12 H\&E and CD3 IHC colorectal cancer slide pairs scanned at $20\times$ (0.454 µm/pixel) with a HAMAMATSU scanner. 
Each pair was prepared with two adjacent slices of the same tumor. The lymphoid infiltration patterns at the same region of adjacent slices are considered to be similar. 
The region of interest (ROI) containing annotations of normal tissue outside the tumor, neoplastic tissue at the tumoral front as well as irrelevant area for lymphoid infiltration assessment were identified by an expert pathologist from Ambroise-Paré Hospital, for both H\&E and IHC slides (Fig.~\ref{fig:pochi}~(A, B)) through the Cytomine \cite{Maree2016} platform.
\par
For lymphocyte segmentation, several datasets were considered in the experiments (training versus validation, see Fig.~\ref{fig:validation}).
SegPath, the largest-scale dataset for the segmentation of cancer histology images, was created using restaining-based annotation approach \cite{DBLP:journals/patterns/KomuraOSOHOHEKIUI23}. 
It consists of 10,453 and 1,082 patches of $984\times984$ pixels scanned at $40\times$ with train and validation splits. 
Each patch comes with its corresponding mask of CD3$^{+}$ and CD20$^{+}$ cells. 
The NuCLS dataset, was created by using collaborative crowd-sourcing annotation pipelines with both medical students and pathologists \cite{DBLP:journals/corr/abs-2102-09099}. 
The corrected single raster version was used. 
470 patches (at 0.2 µm/pixel) containing more than 5 lymphocytes in each were used for validation in this study. 
Finally, the BCa dataset consists of 100 estrogen receptor-positive (ER+) breast cancer (BCa) patches of $100\times100$ pixels scanned at $20\times$, with 3,064 lymphocytes centers identified by an expert pathologist. 
It was released with a lymphocyte detection use case tutorial \cite{JANOWCZYK201629}.
The SegPath dataset will be used for training, while the NuCLS and BCa datasets will be used for external validation.

\section{Method}
\label{sec:methods}
Our methodology (illustrated in Fig.~\ref{fig:pochi}) is applied to evaluate lymphoid infiltration in colorectal cancer (CRC) using in-house standard H\&E stained slides, while ensuring our pipeline's versatility for broader applications.
Sec.~\ref{ssec:segmentation_he} details the lymphocyte segmentation approach tailored for H\&E images. Subsequently, Sec.~\ref{ssec:distance} introduces the distance transform to compute the distances to the tumor margins and Sec~\ref{ssec:infiltration} elaborates on the process of generating lymphoid infiltration curves adjacent to these margins.

\subsection{Lymphocyte segmentation in H\&E}
\label{ssec:segmentation_he}
\begin{figure}[]
 \centering
  \includegraphics[width=0.7\linewidth]{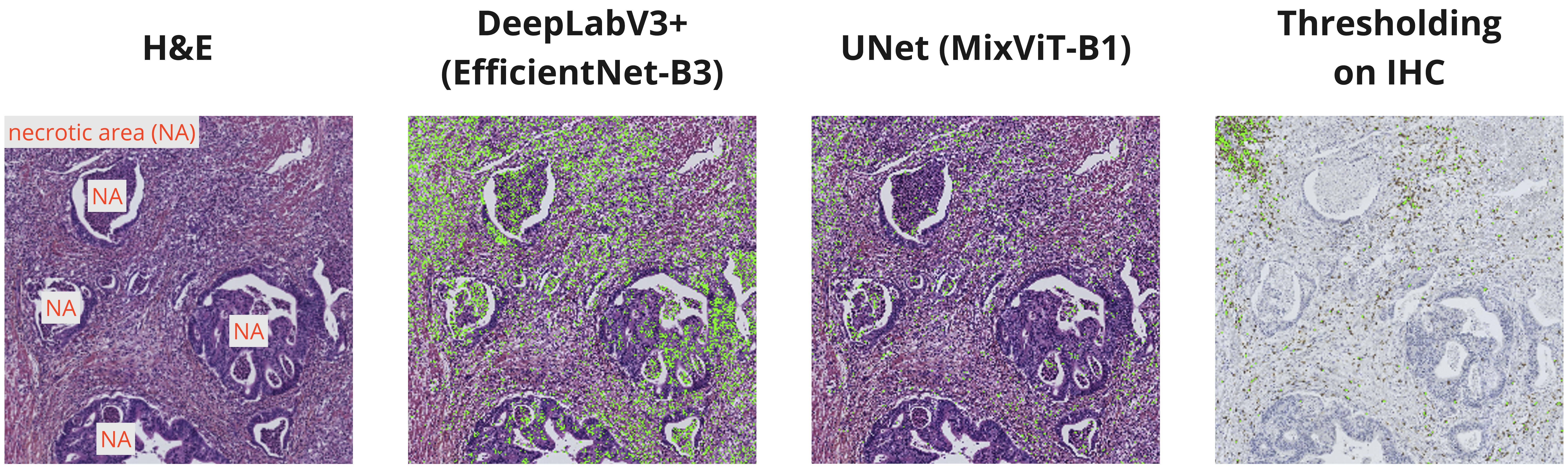}
\caption{Tumor cell in necrotic areas within large tumor glands and lymphocytes appear similar. Using a contextual aware neural network, a U-Net architecture with frozen Mix Vision Transformer encoder (UNet (MixVit-B1)), has better performance in distinguishing necrotic cells and lymphocytes.}
\label{fig:necrotic}
\end{figure}

For H\&E slides, CD3$^{+}$ and CD20$^{+}$ cells are morphologically indistinguishable and it is also not clinically necessary to distinguish them in lymphoid infiltration profiling.
Hence, both CD3$^{+}$ and CD20$^{+}$ cells in H\&E slides are identified (Fig.~\ref{fig:cd3_cd20}). 
Our segmentation strategy relies on training a deep semantic segmentation network on the SegPath dataset to identify the lymphocytes in H\&E slides. 
Although the dimensions of lymphocytes are much smaller compared to the receptive field of FCN, networks with expanded receptive field are of great help in distinguish the cells that are morphologically similar to lymphocytes, by learning contextual information of the surrounding tissues.
For instance, lymphocytes typically have a smooth, rounded nuclear contour with condensed chromatin, giving them a characteristic dark, homogeneous appearance. 
However, the tumor cells in the necrotic areas (Fig.~\ref{fig:necrotic}) also appear with a round shape and a high nucleus-to-cytoplasm ratio, due to cellular degradation and morphological changes. 
In this case, the cellular context and architecture can help distinguish necrotic cells and lymphocytes: the necrotic areas are within large tumor glands.
In this context, we propose to use a modified U-Net \cite{DBLP:conf/miccai/RonnebergerFB15} architecture with a frozen Mix Vision Transformer \cite{DBLP:journals/pr/YuWZG23} encoder initialized with the ImageNet weights.
\par
Color augmentations, random flip / affine transformations and Gaussian blur augmentations were applied in order to help the model generalize to unseen data from different centers, scanned at different magnifications. 
The network was trained on a Nvidia A100 SXM4 80GB GPU for 25 epochs using the
RAdam \cite{DBLP:conf/iclr/LiuJHCLG020} optimizer with a batch size of 16 and an early stopping patience of 5. 
The learning rate was set to 1e-4 and the weight decay rate was set to 1e-4.
\par
The network trained on SegPath was externally validated on NuCLS and BCa datasets (Sec.~\ref{ssec:eval_segmentation}) before applying to the household H\&E slides.

\subsection{Distance transform}
\label{ssec:distance}
For each slide, normal and neoplastic tissue areas within the ROI at the tumoral front, were defined by an expert pathologist (Fig.~\ref{fig:pochi}~(A)).
From these ROIs, lymphocytes are identified (Fig.~\ref{fig:pochi}~(C)) according to the semantic segmentation approach defined above. 
In this study, the focus is the quantification of lymphocyte densities at different distances to the tumor margins. Distance transform can be employed for the distance measurement.
Cellular structures and tissue patterns do not have a preferred orientation in histological images. 
Considering the isotropic nature and the robustness to noise of the Euclidean Distance Transform (EDT), the latter was used for distance measurement in this application. 
The EDT was performed independently for pixels of the normal tissue and of the neoplastic tissue within the ROI (Fig.~\ref{fig:pochi}~(A)). 
In the distance map, each value at the position of the normal tissue stands for the Euclidean distance to its nearest neoplastic tissue (i.e., distance to the tumor margin), vice versa. 
The distances from the normal tissue to the tumor margin were defined as positive, and from the neoplastic tissue were defined as negative. 
The irrelevant area will not contribute to the calculations.
By doing so, the required elements for the lymphoid infiltration curve computation are prepared.

\subsection{Lymphoid infiltration curve}
\label{ssec:infiltration}
From the tumor margin to each side, the lymphocyte pixel densities were calculated within each 10 µm distance step (Fig.~\ref{fig:pochi}~(C)). 
The curves of lymphocyte density against the distance to the tumor margin were then plotted for the ROI of both H\&E and IHC images for comparison purposes (see Sec.~\ref{ssec:eval_curves}). 
Fig.~\ref{fig:pochi}~(E) shows an example of lymphoid infiltration curve from H\&E. 
The abscissa represents the distances to the tumor margin, and the ordinate represents the lymphocyte pixel densities. 
From the infiltration curves, the linear lymphoid infiltration changes across the tumor margins can be qualitatively observed.


\begin{figure}[t]
 \centering
  \begin{minipage}[b]{.4\linewidth}
   \centerline{\includegraphics[width=3.8cm]{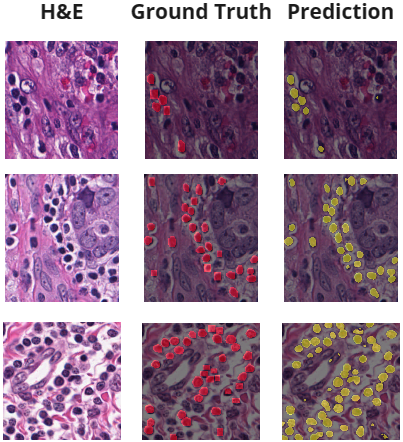}}
   \centerline{NuCLS dataset}\medskip
  \end{minipage}
  ~~
  \begin{minipage}[b]{.4\linewidth}
   \centerline{\includegraphics[width=3.8cm]{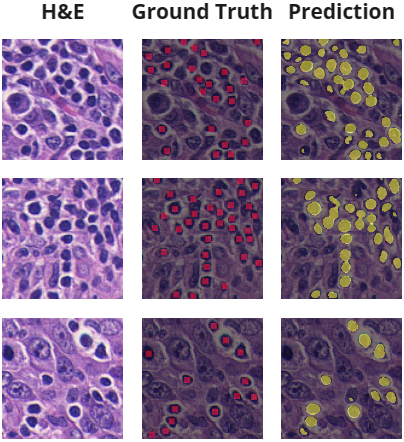}}
   \centerline{BCa dataset}\medskip
  \end{minipage}
\caption{Selected segmentation results on NuCLS and BCa. 
The annotations in the ground truth images of BCa dataset were dilated for visualisation purpose.}
\label{fig:validation}
\end{figure}

\section{Experimental study}
\label{sec:results}
Firstly, the lymphocyte segmentation model used in the proposed lymphoid assessment method was quantitatively evaluated (Sec.~\ref{ssec:eval_segmentation}). 
The similarity of H\&E curves from the proposed method and the IHC curves from the conventional method \cite{DBLP:conf/iciar/DjiroKL18} were then analyzed leading to first validations (Sec.~\ref{ssec:eval_curves}), 
followed by an enhanced clinical study involving a Turing test approach (Sec.~\ref{ssec:turing_test}).

\subsection{Evaluation of lymphocyte segmentation in H\&E}
\label{ssec:eval_segmentation}
During inference, the images from NuCLS and BCa were scaled by $0.9049$ and $2.0$, respectively, to match the scan magnification level of the images in SegPath. 
Similar to the evaluation metric for external validation of the lymphocyte segmentation introduced in \cite{DBLP:journals/patterns/KomuraOSOHOHEKIUI23}, the object-level Dice coefficient was used to evaluate the generalization ability of the trained network to unseen images from different centers / organs, scanned at different magnification levels. 
The patch wise average Dice coefficient recorded for NuCLS dataset is $0.629$ while for BCa dataset is $0.759$. 
The segmentation model used in the proposed method outperforms the two state-of-the-art (SOTA) models \cite{DBLP:journals/patterns/KomuraOSOHOHEKIUI23} trained on SegPath (Table \ref{tab:seg_compare}).
The segmentation accuracy of the trained lymphocyte segmentation neural network shows good generalization ability to different cohorts. 
Some qualitative results are presented in Fig.~\ref{fig:validation}. 

Given the nature of our in-house, label-free lymphocyte cohort, a quantitative analysis of the segmentation performance by the trained model remains unfeasible.
The qualitative comparison suggests that, the contextual aware neural network (i.e., UNet (Mix-Vit-B1)) introduced in Sec.~\ref{sec:methods} has better performance in the scenarios that require the cellular context and architecture to be considered (e.g., distinguishing necrotic cells and lymphocytes within large tumor glands).

\begin{table}[t!]
 \caption{Quantitative (Dice coefficient) comparison of the segmentation methods. 
 The symbol $^{\ast}$ refers to SOTA models trained on SegPath \cite{DBLP:journals/patterns/KomuraOSOHOHEKIUI23}.
 \label{tab:seg_compare}}
 \begin{center}
 \begin{tabular}{|l|l|l|}
 \hline
 Architecture (Encoder) & NuCLS & BCa \\
 \hline
 UNet++ (EfficientNet-B1)$^{\ast}$ & 0.538 & 0.595 \\
 DeepLabV3+ (EfficientNet-B3)$^{\ast}$ & 0.613 & 0.695 \\
 \hdashline
 UNet (MixViT-B1) -- ours & \textbf{0.629} & \textbf{0.759} \\
 \hline
 \end{tabular}
 \end{center}
\end{table}

\begin{figure}[]
 \centering
  \includegraphics[width=0.53\linewidth]{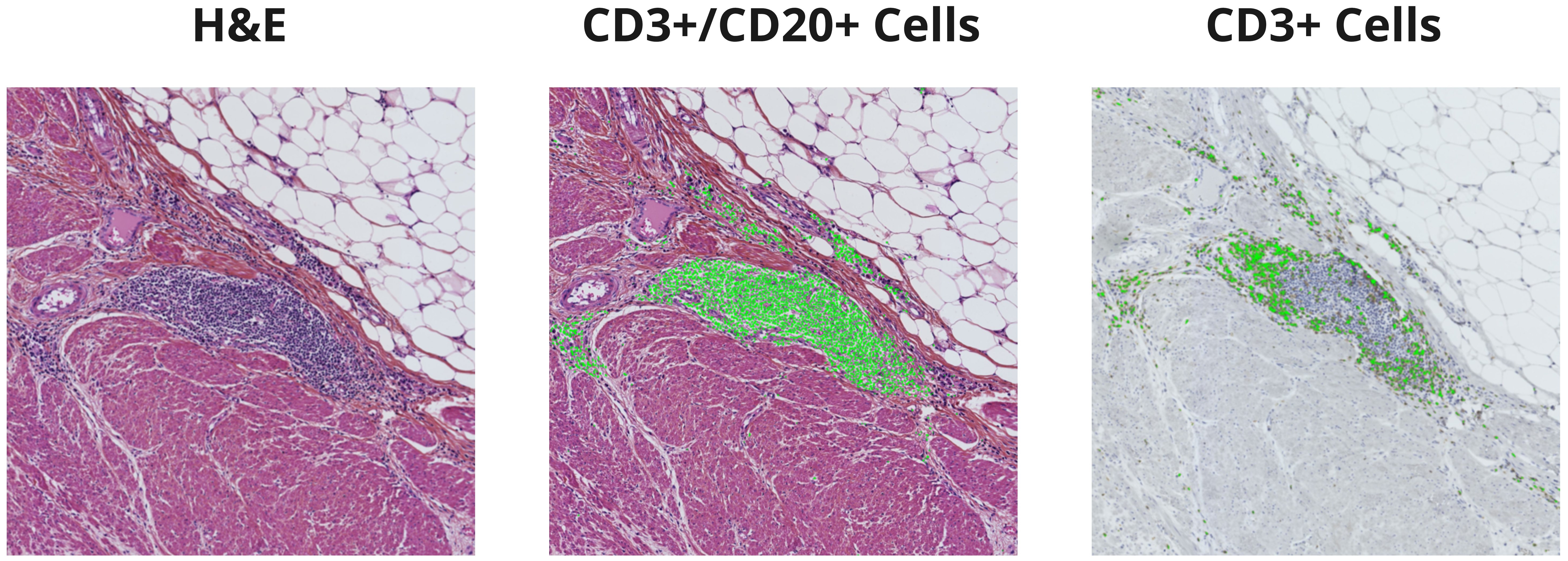}
\caption{A tertiary lymphoid structure (TLS, a lymphoid follicle, a cluster of CD20$^{+}$ cells surrounded by CD3$^{+}$ cells). 
For H\&E images, both CD3$^{+}$ and CD20$^{+}$ will be identified (middle). 
The CD3$^{+}$ in its corresponding IHC lymphocyte cluster (right).}
\label{fig:cd3_cd20}
\end{figure}

\subsection{Evaluation of infiltration curve similarity}
\label{ssec:eval_curves}
Infiltration curves from IHC slides were created as a reference (by following \cite{DBLP:conf/iciar/DjiroKL18}) for similarity comparison of H\&E and IHC curves.
For CD3$^{+}$ cells in IHC images, the stain with diaminobenzidine (DAB) chromogen are isolated through a color deconvolution approach \cite{Ruifrok2001QuantificationOH}, followed by a binary thresholding with a value of $0.095$ (optimal parameter) on the isolated DAB channel, to further eliminate the noise-like structures for refinement (Fig.~\ref{fig:pochi}~(D)). 
IHC lymphocyte infiltration curves are then obtained (Fig.~\ref{fig:pochi}~(F)) using the same process as for H\&E cases.

\par


We approached the comparison of infiltration curves derived from H\&E and IHC stained slides (Fig.~\ref{fig:pochi}~(E, F)) as a quest to uncover temporal series similarities. 
To this end, we examined infiltration curves from 12 colorectal cancer (CRC) pairs, stored in two distinct sets corresponding to each staining method. 
Our objective was to identify blindly, for each H\&E curve, the corresponding IHC curves that exhibited the closest similarity in pattern, thereby establishing a match when both curves originated from the same H\&E-IHC pair.
The constrained Dynamic Time Warping (cDTW) algorithm served as our tool for this analysis, chosen for its effectiveness in accurately measuring similarities between sequences. A shift tolerance of 10 microns was specified to accommodate the nuances of tissue slice annotation. The process included z-score normalization to standardize the series, focusing on the anatomical region from 2.0 mm outside to 2.0 mm inside the tumor margin. For series not covering this specified range, reflection padding was applied. 
From this algorithm, for each H\&E curve, the top-k similar results can then be assesses to retrieve the most similar IHC curves.

\par

Our comparative evaluation, detailed in Table~\ref{tab:curve_compare}, demonstrates that the majority of H\&E-derived infiltration curves find their counterparts within the top-3 matches of the IHC dataset.
Despite this high degree of similarity, perfect alignment between the H\&E and IHC curves is impeded by several factors including:
the derivation of slide pairs from separate tissue slices, the slight variances in manual annotation, the impact of tertiary lymphoid structures (TLS) as illustrated in (Fig.~\ref{fig:cd3_cd20}), and the challenges associated with accurately segmenting lymphocytes on both staining techniques. 
These elements collectively contribute to the occasional discrepancies observed in the alignment of infiltration curves across the two modalities.

\begin{table}[t]
 \caption{H\&E vs. IHC curve matching accuracy: 
 the match rates of H\&E queries to IHC targets are recorded, evaluated within top-k similarity ranks via cDTW. 
 \label{tab:curve_compare}}
 \begin{center}
 \begin{tabular}{|l|l|l|}
 \hline
 Top-k Similarity & cDTW \\
 \hline
 Top-1  & 5/12 \\
 Top-2  & 6/12 \\
 Top-3  & 9/12 \\
 \hline
 \end{tabular}
 \end{center}
\end{table}

\subsection{Enhanced clinical validation: A Turing test approach}
\label{ssec:turing_test}
In assessing the clinical utility of our H\&E-based method for lymphoid infiltration at tumor margins, we employed a Turing test protocol with anonymized curve assessments from H\&E and IHC samples by an expert pathologist. Despite achieving a concordance in 7 out of 12 cases for trial eligibility decisions between the two staining methods, these results underscore the potential of H\&E staining in certain contexts. This partial alignment suggests avenues for refining the approach, underscoring its promise as a supplementary diagnostic tool.

\section{Conclusion}
\label{sec:conclusion}

This study presents a H\&E-based approach for assessing lymphoid infiltration, offering a viable alternative to traditional IHC methods. Employing a deep learning network trained on an extensive public dataset, we achieved precise segmentation of CD3$^{+}$ and CD20$^{+}$ cells, facilitating a thorough quantitative analysis.
The clarity and stability provided by H\&E staining in tumor margin delineation, less affected by tissue preservation variances, reinforce its suitability for immunotherapy pre-screening applications. 
Our application of the Turing test advocates for a paradigm shift in validation methods that reflect clinical practice, supported by our medical team as a promising approach that aligns with real-world decision-making.
This work not only advances the integration of lymphoid infiltration assessment into clinical practice but also lays the groundwork for future enhancements, including the exploration of specific cut-off values for identifying patients with high immune infiltration and automating the segmentation of neoplastic tissue.

%
%

\bibliographystyle{splncs04}
\bibliography{ref}
\end{document}